# Assessing Text Classification Methods for Cyberbullying Detection on Social Media Platforms

Adamu Gaston Philipo , Doreen Sebastian Sarwatt , Jianguo Ding , *Member, IEEE,* Mahmoud Daneshmand , *Member, IEEE,* Huansheng Ning , *Member, IEEE*

.

*Abstract*—**Cyberbullying significantly contributes to mental health issues in communities by negatively impacting the psychology of victims. It is a prevalent problem on social media platforms, necessitating effective, real-time detection and monitoring systems to identify harmful messages. However, current cyberbullying detection systems face challenges related to performance, dataset quality, time efficiency, and computational costs. This research aims to conduct a comparative study by adapting and evaluating existing text classification techniques within the cyberbullying detection domain. The study specifically evaluates the effectiveness and performance of these techniques in identifying cyberbullying instances on social media platforms. It focuses on leveraging and assessing large language models, including BERT, RoBERTa, XLNet, DistilBERT, and GPT-2.0, for their suitability in this domain. The results show that BERT strikes a balance between performance, time efficiency, and computational resources: Accuracy of 95%, Precision of 95%, Recall of 95%, F1 Score of 95%, Error Rate of 5%, Inference Time of 0.053 seconds, RAM Usage of 35.28 MB, CPU/GPU Usage of 0.4%, and Energy Consumption of 0.000263 kWh. The findings demonstrate that generative AI models, while powerful, do not consistently outperform fine-tuned models on the tested benchmarks. However, state-of-the-art performance can still be achieved through strategic adaptation and fine-tuning of existing models for specific datasets and tasks.**

## I. INTRODUCTION

THE Cyberbullying.[1] has become a major issue in digital spaces, significantly affecting mental health, especially among adolescents. The anonymity and lack of restraint provided by online platforms enable harmful behavior, leading to a rise in cyberbullying incidents that can occur at any time and in any place [1]. Victims frequently suffer from serious mental health problems, such as anxiety, depression, and even suicidal thoughts [2].

Over 56% of adolescents report cyberbullying experiences [3]. Victims face higher risks of anxiety, stress, and low self-esteem, with some considering suicide [2]. Online anonymity

H. Ning, A. Philipo, and D. Sarwatt are with the School of Computer and Communication Engineering, University of Science and Technology Beijing, Beijing, China. Corresponding author: Huansheng Ning (e-mail: ninghuansheng@ustb.edu.cn).

J. Ding is with the Department of Computer Science, Blekinge Institute of Technology, 371 79, Karlskrona, Sweden.

M. Daneshmand with the Department of Business Intelligence and Analytics and the Department of Computer Science, Stevens Institute of Technology, Hoboken, NJ, USA fuels cyberbullying, disproportionately affecting girls [1]. Addressing cyberbullying requires collaboration among educators, parents, and policymakers to support victims and curb harmful behaviors [4]. Although cyberbullying's negative effects are emphasized, some argue that digital platforms can also offer supportive communities to help victims cope. This duality highlights the complex impact of online interactions on mental health.

[1] Cyberbullying is a form of bullying that occurs through digital platforms, such as social media, messaging apps, gaming platforms, or other online forums.

Cyberbullying detection has progressed significantly, leveraging various machine learning (ML) and deep learning (DL) methods. However, critical challenges still affect their effectiveness. Traditional ML models like Random Forest, SVM, and Naive Bayes are commonly used but often fail to interpret the nuanced language and context of social media posts [5], [6]. Advanced models, including CNN, RNN, and BERT, show better accuracy, with BERT reaching up to 88.8% in binary classification tasks [5]. However, these models need extensive labeled datasets, which are often scarce [7]. Many datasets focus mainly on aggressive language, overlooking intent and peer dynamics, which are essential for comprehensive detection [8]. Current models also struggle to fully capture interaction context, leading to false positives or negatives [6], [9]. Despite advancements, challenges remain in dataset diversity and contextual understanding, highlighting the need for further research to improve accuracy and reliability.

Enhancing the performance and speed of cyberbullying detection is critical for mitigating the impacts of online harassment. Efficient detection systems allow timely intervention, helping protect victims and foster safer online environments. Multi-algorithmic methods, such as combining Support Vector Classifier (SVC) with TF-IDF, have shown improved accuracy over single algorithms in identifying cyberbullying [10]. Transformer models, especially DistilBERT, have achieved notable results, with 94.36% accuracy and 93.91% recall, demonstrating the power of advanced Natural Language Processing (NLP) techniques [11]. Additionally, stacking ensemble learning has reduced detection time to just 3 minutes while sustaining a high accuracy of 97.4%, enabling prompt intervention that can prevent escalation [12]. Integrating user interaction dynamics and temporal coherence into detection models can further enhance predictive accuracy by capturing the repetitive patterns of bullying behavior [13].

Datasets of cyberbullying also become one of the challenges because the data needs to be collected, preprocessed, and labeled manually. These tasks are costly and time-consuming, requiring significant resources to accomplish [14]. Although there are semi-automatic and automatic methods for labeling data, they remain insufficient compared to the manual approach because manual labeling enhances contextual understanding in the relevant language. Consistency during the annotation process is crucial at all stages, including class distribution [15]. It is important to establish a common understanding of cyberbullying forms during the labeling process. Here, Inter-Annotator Agreement (IAA) plays a critical role in maintaining consistency across various forms of cyberbullying, such as hate speech, offensive language, insults, and sexual harassment. The main challenge lies in the fact that most cyberbullying datasets contain diverse forms of cyberbullying, making it difficult to achieve contextual understanding and comparison due to inconsistencies between them [16]. Inter-Annotator Agreement (IAA) can help mitigate this issue, highlighting the significance of this study in comparing cyberbullying datasets used for detection on social media platforms.

Computational resource usage during the evaluation process is another challenge, particularly in cyberbullying detection. This issue is influenced by both the technologies used and the size and characteristics of the cyberbullying datasets. Computational resources include RAM usage, CPU or GPU utilization, and energy consumption during inference [17]. Developing solutions that require excessive computational resources can be counterproductive. Modern technologies aim to provide solutions with high performance, real-time applicability, and minimal energy consumption. Energy efficiency is especially important in developing countries where resource constraints are a major concern. This challenge highlights the importance of this study in comparing detection methods based on computational resource requirements.

Time efficiency is yet another challenge in cyberbullying detection methods on social media platforms. This issue is influenced by both the technologies employed and the size of the datasets. Some detection methods have multiple layers, resulting in prolonged loops for data processing, especially during the evaluation phase [18]. Additionally, large datasets can significantly increase processing time. These challenges necessitate the identification of the best detection methods that minimize inference time while accurately predicting instances of cyberbullying on social media platforms. This aspect is particularly significant for real-time applications, where continuous monitoring is required to identify harmful text in seconds, thereby mitigating the impact of cyberbullying on communities.

### A. Contributions

The key contributions of this study are outlined as follows:

1) This work is the first to systematically conduct a comprehensive comparative study of large language models (LLMs) for detecting cyberbullying. The study spans multiple social media platforms to evaluate the adaptability and effectiveness of these techniques. Philipo et al. [9] conducted comparative study on machine and deep learning model in previous study.

2) Datasets are analyzed based on their source, size, scope, diversity, generalizability, and annotation quality, as these factors are crucial for cyberbullying detection on social media platforms.

3) Existing general text classification techniques are adapted and fine-tuned specifically for the cyberbullying detection domain, assessing their performance across various social media platforms.

4) The time efficiency of the adapted detection techniques is measured by analyzing computational costs, detection and error rates, and potential for real-time application. This analysis identifies approaches that offer both high performance and practical viability in real-world scenarios.

### B. Paper Organization

This paper is organized as follows: Section 2 reviews the related work, while Section 3 presents the detection methods and evaluation metrics. Section 4 outlines the datasets and experimental setup, including a detailed analysis. Section 5 provides a thorough discussion of the results. Finally, Section 6 offers a comprehensive future directions and conclusion of the research.

## II. RELATED WORK

This section offers a comprehensive overview of existing research and introductory knowledge on cyberbullying detection methods within the domain of cyberbullying classification. It aims to establish a clear understanding of the challenges and requirements involved in applying detection techniques to text classification under cyberbullying scenarios. Additionally, it explores various approaches used for cyberbullying detection, highlighting current advancements, strengths, weaknesses, and research gaps in this field. The following are related works conducted so far in the domain of cyberbullying detection across different online communication platforms.

Khan et al. [19] proposed a deep sequence model, BLSTM, for detecting offensive language in Roman Pashto. This model was compared to traditional machine learning models, including NB, LR, SVM, and RF. The BLSTM model was trained and tested using three feature extraction methods: BoW, TF-IDF, and sequence integer encoding. Results indicate that BLSTM outperformed the other machine learning models, achieving an accuracy of 97.21%.

Bauer et al. [20] expanded on GPT benchmarking by converting GPT models into classifiers and testing them on three distinct tasks: hate speech detection, offensive language detection, and emotion classification using Twitter datasets. Zero-shot and few-shot approaches were applied to assess the classification abilities of the GPT models. The results indicate that GPT models do not always surpass fine-tuned models on these benchmarks. However, using GPT-3 with a few-shot approach achieves state-of-the-art performance in hate speech and emotion detection, and GPT-4 shows increased sensitivity to the examples provided in the few-shot setup.

Obaida et al. [21] proposed a deep learning to detect instances of cyberbullying on social media platforms. The experiments were conducted using three datasets from Twitter,

Instagram, and Facebook, leveraging Long Short-Term Memory (LSTM) for prediction. The model achieved accuracies of approximately 96.64%, 94.49%, and 91.26% for the Twitter, Instagram, and Facebook datasets, respectively.

Alabdulwahab et al. [22] proposed various machine learning and deep learning algorithms, including KNN, NB, DT, SVM, RF, and LSTM to detect instances of cyberbullying on social media platforms. Using a Twitter dataset in the experiment, the LSTM model outperformed the other models, achieving an accuracy of 96%, compared to 90% for KNN and 92% for SVM.

Mehendale et al. [23] designed and developed an effective technique for detecting online abusive and bullying messages by combining Natural Language Processing (NLP) and machine learning. This model can identify offensive or hateful language in both English and Hinglish. In the experiment, TF-IDF and CountVectorizer were used for feature extraction. The machine learning models employed included SVM, LR, RT, KNN, DT, Bagging, SG, Adaboost, and NB. The results showed that Random Forest outperformed the other machine learning models, achieving an accuracy of 97.1% when using TF-IDF.

Raj et al. [24] proposed a deep learning framework (CNN-BiLSTM architecture) designed to evaluate real-time tweets or social media posts and accurately identify cyberbullying content. The application can detect cyberbullying in multilingual data, including English, Hindi, and Hinglish. The results show that the CNN-BiLSTM architecture outperformed other deep learning models, achieving an accuracy of 95%.

Roy and Mali [25] developed a deep transfer model (2DCNN) to address image-based cyberbullying on social platforms such as Facebook, Twitter, and others. Initially, a convolutional neural network (CNN) was used for model development, followed by the application of transfer learning models. The proposed model achieved an accuracy of 89% in the best case, demonstrating its ability to effectively detect most cyberbullying posts.

Chiu et al. [26] proposed using the large language model OpenAI's GPT-3.0 to identify hate speech and classify it as either sexist or racist for zero-shot, one-shot, and few-shot learning approaches. The results show that the few-shot learning approach outperformed the other learning approaches, achieving an accuracy of 85%.

Azeez et al. [27] conducted research on cyberbullying detection on social media using an artificial intelligence approach. Several classification algorithms were reviewed and evaluated to find the most suitable one for detecting cyberbullying. A new ensemble model was proposed to identify cyberbullying content using English Twitter datasets, labeled as either cyberbullying or non-cyberbullying. Ten classifiers were tested, including K-Nearest Neighbors, Logistic Regression, Naive Bayes, Decision Tree, Random Forest, Adaptive Boosting, Stochastic Gradient Descent, Linear Support Vector Classifier, and Bagging, along with an ensemble classifier. The ensemble classifier produced the best results, while the Linear Support Vector Classifier had the lowest performance. Random Forest was one of the top classifiers, achieving 77% accuracy, 73% precision, and 94% recall. The ensemble model improved performance, with averages of 77% accuracy, 66% precision, and 94% recall, compared to the Linear Support Vector Classifier's 59%, 42%, and 86%.

Gutha et al. [28] presented deep learning approaches involving Long Short-Term Memory (LSTM) networks combined with Convolutional Neural Networks (CNN) and pre-trained models based on Bidirectional Encoder Representations from Transformers (BERT), such as IndicBERT. Hate speech detection was performed across three languages: Bengali, Bodo, and Assamese. The results highlight the effectiveness of these approaches, with IndicBERT achieving an impressive F1 score of 69.73% for Assamese, MuRIL reaching 71.96% for Bengali, and a BiLSTM model enhanced by an additional Dense layer attaining 83.51% for Bodo.

Table I provides a summary of existing systems. While research has shown promising results, further improvement is needed in model effectiveness, generalizability, and scalability. Advancing these models with ongoing development and advanced techniques can lead to reliable, scalable tools that help mitigate cyberbullying's harmful effects.

This research aims to bridge the gap by adapting and evaluating detection methods from the text classification domain for their effectiveness in identifying cyberbullying text. The study uses five key datasets and eight detection methods to improve detection capabilities. It also examines the new factor of detection time, focusing on the importance of quick and real-time detection to handle cyberbullying efficiently.

## III. DETECTION METHODS AND EVALUATION METRICS

This section examines five detection methods from text classification, analyzing their suitability for identifying cyberbullying on social media platforms. These methods were chosen because they have shown success in text classification tasks. The evaluation is performed systematically across different datasets, models, and scenarios related to cyberbullying to provide a complete analysis. The aim is to classify harmful text after detecting cyberbullying, as shown in Fig.1. Each method's setup, including its configurations, hyperparameters, and technical details, is explored. Their performance and efficiency are evaluated using well-defined criteria.

### A. Detection Methods

1) BERT [29] is a powerful model in natural language processing (NLP). It uses a transformer-based architecture to achieve excellent results in text classification tasks. BERT's key feature is its ability to understand context in both directions (deeply bidirectional) and its unique pretraining process, which involves predicting masked words in text. This allows BERT to capture complex relationships and meanings in sentences effectively, even in challenging contexts. In this study, BERT was fine-tuned for detecting cyberbullying on social media platforms. Key parameters, such as epochs, batch size, and learning rate, were adjusted during training. BERT utilizes only the encoder stack of the Transformer architecture. It employs Masked Language Modeling (MLM), where tokens are randomly masked and then predicted, along with Next Sentence Prediction (NSP), which determines whether two sentences are sequential. Additionally, it combines token, segment, and positional embeddings to create its input representation. Mathematical expression of BERT model is as follows:

$$L_{MLM} = -\sum_{i \in \text{masked}} \log P(x_i | \text{context}) \qquad (1)$$

$$L_{NSP} = -y \log P(\text{is\_next}) + (1-y) \log P(\text{not\_next}) \qquad (2)$$

TABLE I
SUMMARY OF RELATED WORK

| Title | Year | Strengths | Weaknesses | Research Gap |
|-------|------|-----------|------------|--------------|
| Offensive Language Detection for Low Resource Language Using Deep Sequence Model [19] | 2024 | A BiLSTM model achieved an accuracy of 97.21% for offensive language detection in the Roman Pashto language. | Computationally expensive | How to design and develop a deep sequence model with the highest performance and lower computational complexity. |
| Offensiveness, Hate, Emotion and GPT: Benchmarking GPT3.5 and GPT4 as Classifiers on Twitter-specific Datasets [20] | 2024 | GPT models can be used as classifiers in hate speech detection, offensive language detection, and emotion classification, achieving reasonable performance with F1 scores of 69.9%, 80.5%, and 82.2%, respectively. | Expensive and unable to verify if test set contamination was involved, GPT models do not always outperform fine-tuned models on these benchmarks. | How to enhance the performance of GPT models to reach the same level as, or exceed, fine-tuned models. |
| Deep Learning Algorithms for Cyber-Bulling Detection in Social Media Platforms [21] | 2024 | The LSTM model achieved accuracies of 96.64%, 94.49%, and 91.26% for the Twitter, Instagram, and Facebook datasets, respectively. | Small size dataset. | How to gather a more extensive dataset. |
| Cyberbullying Detection using Machine Learning and Deep Learning [22] | 2023 | CNN+LSTM achieved an accuracy of 96% in detecting cyberbullying tweets, compared to ensemble models: KNN with an accuracy of 90% and SVM with an accuracy of 92%. | In KNN, it is required to update bad words periodically, and it requires a large amount of memory to store the group of words. SVM is not suitable for large datasets, as it is very expensive and consumes a lot of energy. | How to design a model to detect cyberbullying tweets with minimal computational cost. |
| Cyber bullying detection for Hindi-English language using machine learning [23] | 2022 | The Ensemble (Random Forest) model achieved an accuracy of 97.1% in detecting cyberbullying in the Hindi-English language. | Scarcity of labeled larger dataset, Manual feature extraction, time inefficiency, imbalance of classes. | How to gather a more labeled dataset, how to automatically integrate feature extraction, how to improve time efficiency by reducing complexity and how to address bias data. |
| An Application to Detect Cyberbullying Using Machine Learning and Deep Learning Techniques [24] | 2022 | The CNN+BiLSTM model achieved an accuracy of 95% in detecting cyberbullying tweets in multilingual data. | Language limitations, narrow scope, and absence of contextual comprehension | How to enhance the accuracy of the model for detecting cyberbullying text in multiple languages. |
| Cyberbullying detection using deep transfer learning [25] | 2022 | The 2D CNN model achieved an accuracy of 89% in detecting cyberbullying in the MMHS dataset. | Does not consider textual cyberbullying detection and does not consider text-image combinations in cyberbullying posts. | How to improve the model to detect textual cyberbullying and text-image combinations in cyberbullying posts. |
| Detecting Hate Speech with GPT-3 [26] | 2022 | OpenAI's GPT-3 Text-davinci-001 achieved an accuracy of 85% with few-shot learning to classify cyberbullying tweets in the ETHOS dataset. | Expensive, small-sized datasets, dataset imbalance, limited control, and overfitting. | How to gather a more extensive and balanced class dataset to avoid overfitting. |
| Cyberbullying Detection in Social Networks: Artificial Intelligence Approach [27] | 2021 | The ensemble model (SVC, LR, NB) achieved an accuracy of 94% in detecting cyberbullying tweets. | The dataset discloses information about users, while fields such as age and gender of posters are unavailable. It is also limited to English tweets and data | How to develop a multilingual model to detect cyberbullying tweets, how to gather a more extensive dataset that includes posters |
| Multilingual Hate Speech and Offensive Language Detection of Low Resource Languages [28] | 2021 | IndicBERT achieved an F1 score of 69.726%, MuRIL achieved an F1 score of 71.955%, and BiLSTM achieved an F1 score of 83.513% for Assamese, Bengali, and Bodo, respectively. | Small-sized dataset, Bodo lacks dedicated pre-trained models, leading to overfitting. | How to gather a more extensive dataset to avoid overfitting. |

2) RoBERTa [30] is a state-of-the-art language model designed for natural language processing (NLP) tasks. It enhances performance across various NLP applications by leveraging a larger training dataset and an optimized training process. RoBERTa has been successfully employed in numerous domains, including the detection of cyberbullying on social media platforms. RoBERTa improves upon BERT by removing the Next Sentence Prediction (NSP) task and introducing dynamic masking, where the masked tokens change at each epoch. It is pre-trained using larger batches and more extensive datasets while retaining the same embedding structure as BERT. The model focuses exclusively on Masked Language Modeling (MLM), optimizing it with more robust training techniques. Mathematical expression of RoBERTa model is as follows:

$$L_{MLM} = -\sum_{i \in \text{masked}} \log P(x_i | \text{context}) \qquad (3)$$

Masking is applied dynamically across training epochs, meaning $x_i$ changes at every iteration.

3) XLNet [31] is a state-of-the-art large language model that enhances natural language processing (NLP) tasks through its innovative Permutation Language Modeling (PLM) approach. This model integrates the strengths of both autoencoding and autoregressive methods, allowing it to capture long-range dependencies and contextual nuances effectively. XLNet has demonstrated remarkable performance across various applications, including cyberbullying classification, sentiment analysis, personality recognition, and argumentation element annotation. XLNet combines the strengths of BERT and autoregressive models like GPT by utilizing Permutation Language Modeling (PLM), which predicts tokens in a random order. This approach enables it to capture bidirectional context while maintaining autoregressive properties. Its embedding structure is similar to BERT but incorporates positional embeddings differently to accommodate the permutations. Mathematical expression of XLNet model is as follows:

$$L_{PLM} = -\sum_{t=1}^{T} \log P(x_t | x_{<t}, \pi) \qquad (4)$$

where $\pi$ is a permutation of the input sequence, and $x_{<t}$ represents the context tokens preceding $x_t$ in the permuted order.

4) DistilBERT [32], a distilled version of the BERT (Bidirectional Encoder Representations from Transformers) model, is designed to retain BERT's performance while significantly reducing its size and computational demands. This streamlined model has become popular across various applications, showcasing its efficiency and adaptability in natural language processing (NLP) tasks such as text classification. DistilBERT reduces BERT's size to enhance efficiency by employing knowledge distillation, where the smaller model is trained to replicate the behavior of the larger BERT model. It eliminates the Next Sentence Prediction (NSP) objective and reduces the embedding size and the number of attention heads. Mathematical expression of DistilBERT model is as follows:

$$L = \alpha \cdot L_{\text{teacher}} + \beta \cdot L_{\text{student}} \qquad (5)$$

where $L_{\text{teacher}}$ represents the loss from the larger BERT model, and $L_{\text{student}}$ is the loss from the smaller, distilled model.

5) GPT-2.0 [33] is a large language model that is part of the transformer model family, which has transformed tasks in natural language processing (NLP) like text generation, translation, summarization, and answering questions. In this study, GPT-2.0 was adapted into a classifier for text classification to detect cyberbullying on social media platforms. GPT-2 utilizes only the decoder stack of the Transformer architecture and is designed for Causal Language Modeling, predicting the next token in a sequence. It focuses on generating coherent and contextually rich text by combining token embeddings with positional embeddings. Mathematical expression of GPT-2.0 model is as follows:

$$L_{CLM} = -\sum_{t=1}^{T} \log P(x_t | x_{<t}) \qquad (6)$$

All the aforementioned detection methods are built upon the Transformer architecture, specifically leveraging its encoder or decoder blocks. Transformers utilize a Self-Attention Mechanism to compute relationships between all tokens in the input sequence, as expressed mathematically below:

$$\text{Attention}(Q, K, V) = \text{softmax}\left(\frac{QK^T}{\sqrt{d_k}}\right)V \qquad (7)$$

where Q, K, and V are query, key, and value matrices derived from input embeddings, and $d_k$ is the dimensionality.

The technical differences among these detection methods can be categorized into key features such as architecture, objectives, training data, pretraining tasks, fine-tuning, efficiency, and contextualization.

In terms of architecture, BERT, RoBERTa, and DistilBERT use an Encoder-only model, while XLNet uses both Encoder-only and Autoregressive, and GPT-2.0 uses a Decoder-only model. Regarding objectives, BERT aims for Masked Language Modeling (MLM) and Next Sentence Prediction (NSP), RoBERTa focuses on MLM, XLNet targets Permutation Language Modeling (PLM), DistilBERT is a distilled version of MLM, and GPT-2.0 is designed for Causal Language Modeling. For training data, BERT requires large corpora, RoBERTa needs even larger corpora, XLNet requires similar data to BERT but enhanced, DistilBERT uses a subset of BERT's corpora, and GPT-2.0 demands vast datasets like WebText. In pretraining tasks, BERT employs a bidirectional approach, RoBERTa utilizes a robust MLM, XLNet applies a permuted bidirectional model, DistilBERT uses an efficient bidirectional method, and GPT-2.0 adopts a unidirectional approach. Regarding fine-tuning, BERT is fine-tuned for task-specific purposes such as Question and Answering, RoBERTa and XLNet follow similar approaches to BERT, DistilBERT is fine-tuned in the same way as BERT, while GPT-2.0 is suited for few-shot or zero-shot tasks. In terms of efficiency, BERT and RoBERTa perform moderately, XLNet is highly efficient but more complex, DistilBERT offers high efficiency with a lightweight model, and GPT-2.0 performs moderately. Lastly, for contextualization, BERT and RoBERTa utilize full bidirectional context, XLNet uses a permuted bidirectional context, DistilBERT follows the same approach as BERT, and GPT-2.0 employs unidirectional context [34], [35], [36].

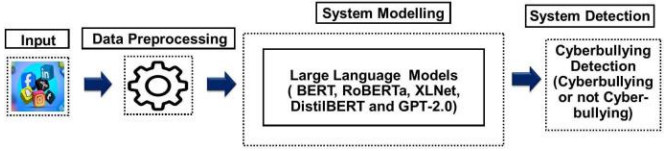

Fig. 1. Detection Methods.

### B. Evaluation Metrics

Six evaluation metrics were used to assess the performance and improve the time efficiency of large language models (LLMs) in detecting cyberbullying on social media platforms: accuracy score, precision, recall, F1 score, error rate (e), and inference time (IT). Additionally, computational resource usage was evaluated through RAM usage, CPU/GPU usage, and energy consumption to identify the detection method that requires the least computational resources. These metrics provide a comprehensive evaluation of the algorithms' ability to accurately differentiate between cyberbullying and non-cyberbullying texts.

Table II presents a confusion matrix that summarizes the performance of a classification model by comparing predicted and actual classifications for each data point. The diagonal entries show correct predictions, while off-diagonal entries indicate errors. In this matrix: True Positive (TP) represents texts correctly identified as cyberbullying, True Negative (TN) indicates texts correctly identified as non-cyberbullying, False Positive (FP) counts texts incorrectly classified as cyberbullying, and False Negative (FN) counts texts incorrectly classified as non-cyberbullying.

1) The accuracy score is determined by the percentage of texts in the dataset that were correctly classified. This is how the accuracy is calculated:

$$\text{Accuracy} = \frac{TP + TN}{TP + FP + FN + TN} \quad (8)$$

2) Precision represents the ratio of correct positive predictions to the total number of positive predictions made. The formula for calculating precision is as follows:

$$\text{Precision} = \frac{TP}{TP + FP} \quad (9)$$

3) Recall is the ratio of actual positive instances that are correctly identified as positive. The formula for calculating recall is as follows:

$$\text{Recall} = \frac{TP}{TP + FN} \quad (10)$$

4) F1 Score: The F1 score is a metric derived by calculating the harmonic mean of precision and recall. The formula for the F1 score is as follows:

$$\text{F1 Score} = 2 \times \frac{Precision \times Recall}{Precision + Recall} \quad (11)$$

5) Error Rate (e): This is the percentage of misclassified instances. It can be calculated as:

$$\text{Error Rate (e)} = \frac{FP + FN}{N} \quad (12)$$

where N represents the total number of instances used during prediction, which is split from the original dataset. A lower error rate indicates an overperforming and more effective method for detecting cyberbullying on social media platforms.

6) Inference Time ($T_{start}$) is the average time taken by the model to make predictions on a single cyberbullying instance. The formula for the inference time is as follows:

$$t_{instance} = \frac{T_{end} - T_{start}}{N} \quad (13)$$

where N represents the total number of instances used during prediction, which is split from the original dataset, $T_{start}$ is the is the start time for the prediction process, and $T_{end}$ is the end time for the prediction process. A shorter inference time indicates a faster and more effective method for detecting cyberbullying on social media platforms.

TABLE II
Confusion Matrix for Cyberbullying Detection

| Actual \Predicted | Positive | Negative |
|---|---|---|
| **Positive** | TP (True Positive) | FP (False Positive) |
| **Negative** | FN (False Negative) | TN (True Negative) |

## IV. Experiment Setup

### A. Datasets

This study on cyberbullying detection on social media platforms utilizes datasets from three sources: Kaggle, IEEE Data Port, and Tweeteval. The Kaggle dataset contains 47,693 tweets, the IEEE Data Port dataset includes 2,141 comments, and the Tweeteval dataset holds 51,987 comments as summarized in the Table III.

*1) Kaggle Dataset:* This dataset consists of texts related to cyberbullying, collected primarily from Twitter and YouTube. The data is categorized into five classes: not cyberbullying, age, gender, religion, and ethnicity. The dataset is structured with two primary columns to categorize and identify instances of cyberbullying: 'tweet text' (the tweet content) and 'label' (cyberbullying category).

*2) IEEE Data Port Dataset:* This dataset primarily features texts gathered from Twitter due to its ease of accessibility, along with smaller contributions from Instagram and Facebook. The data is divided into five classes: slut-shaming, sexual harassment, revenge porn, cyberstalking, and doxing. The dataset is structured with two primary columns to categorize and identify instances of cyberbullying: 'tweet' (comment text) and 'class' (cyberbullying category).

TABLE III
SUMMARY OF CYBERBULLYING DATASETS

| Dataset | Kaggle [37] | | | | | IEEE Data Port [38] | | | | | Tweeteval [ ] | | | | |
|---|---|---|---|---|---|---|---|---|---|---|---|---|---|---|---|
| | not cyberbullying | Gender | Ethnicity | Religion | Age | Slut Shaming | Sexual Harassment | Cyberstalking | Revenge Porn | Doxing | Hate | Offensive | Emotion | Religion | Spiritual |
| Parameter (%) | 19.90 | 20.00 | 20.00 | 20.10 | 20.00 | 23.40 | 23.40 | 14.20 | 18.50 | 20.60 | 24.80 | 27.10 | 9.70 | 20.20 | 18.20 |
| Generated Year | 2020 | | | | | 2021 | | | | | 2020 | | | | |
| Size | 47,693 | | | | | 2,141 | | | | | 51,987 | | | | |

*3) TweetEval Dataset:* This dataset contains texts sourced from Twitter and is organized into five classes: hate, offensive, emotion, religion, and spiritual. The dataset is structured with two primary columns to categorize and identify instances of cyberbullying: 'text' (comment content) and 'label' (cyberbullying type).

The substantial size of these datasets makes them highly suitable for training and testing large language models (LLMs) for cyberbullying detection on social media platforms. The extensive data allows for robust model training, enhancing the model's ability to generalize to new, unseen data. Additionally, the volume of the datasets enables an in-depth analysis of cyberbullying trends and patterns across social media platforms.

### B. Hardware Configuration used in Experiments

The details of the hardware configuration used in this study are summarized in Table IV. The setup was designed to maximize performance and support the efficient execution of large language models (LLMs) utilized in the experiments.

TABLE IV
HARDWARE CONFIGURATION USED IN EXPERIMENTS

| Component | Specifications |
|---|---|
| CPU | Intel® Core™ i9-13900, 3.60 GHz |
| RAM | 32 GB |
| GPU | NVIDIA GeForce RTX 4090, 12 GB |
| Operating System | Windows 11 Pro (64-bit) |
| Programming Language | Python 3.9 |
| Framework | Keras with TensorFlow 2.5 |

### C. Baseline Classifiers

This study employed five text classifiers: BERT, RoBERTa, XLNet, DistilBERT, and GPT-2.0 to classify cyberbullying instances on social media platforms using three distinct datasets: Kaggle, IEEE Data Port, and TweetEval.

All classifiers were trained using the ReLU activation function and cross-entropy loss. The Adam optimizer was employed during training, with a learning rate of 0.0005. An early stopping mechanism was implemented to prevent overfitting by halting training when no improvement in results was observed. The training process utilized a fixed batch size of 32 and was conducted over 10 epochs for each dataset, with an 80:20 split between training and testing data. This methodology aimed to demonstrate the adaptability of the detection techniques, emphasizing that their effectiveness is not limited to a specific classifier. The findings confirmed the broad applicability of these approaches for text classification within the domain of cyberbullying detection.

### D. Detection Methods Implementation

The detection techniques were applied exactly as described by their original authors to guarantee their correct implementation. Each method was set up based on the instructions in the initial research. By following this procedure, the evaluation remains unbiased and maintains its accuracy, enabling a thorough and reliable comparison of the performance of each method in identifying cyberbullying in social media text classification.

## V. RESULTS AND DISCUSSION

This section analyzes five detection methods: BERT, RoBERTa, XLNet, DistilBERT, and GPT-2.0 across three datasets. The evaluation focuses on key metrics: Accuracy, Precision, Recall, F1 Score, Error Rate, Inference Time, RAM Usage, CPU/GPU Utilization, and Energy Consumption during inference, as detailed in Table V and VII. Visual summaries of these metrics for the various detection methods are presented in Fig.2 to 10. This comprehensive evaluation provides valuable insights for enhancing the safety and reliability of social media platforms through effective cyberbullying detection methods.

### A. Comparison of Accuracy

*1) Kaggle Dataset:* BERT achieved the highest accuracy (94%) among the methods, followed closely by RoBERTa and XLNet, both at 93%, while DistilBERT also performed well with an accuracy of 92%. In contrast, GPT-2.0 showed the lowest performance on this dataset with 85% accuracy. These results highlight that BERT-based models, particularly BERT, RoBERTa, and XLNet, significantly outperform GPT-2.0, suggesting that fine-tuned transformer-based models designed for classification tasks, such as BERT and RoBERTa, are more effective than generative models like GPT-2.0.

*2) IEEE Data Port Dataset:* DistilBERT achieved the highest accuracy (87%) on this dataset, outperforming other methods, followed closely by XLNet with an accuracy of 85%. BERT achieved 83% accuracy, while RoBERTa performed slightly lower at 82%, and GPT-2.0 showed the lowest performance at 76%. DistilBERT's strong performance demonstrates that lighter, more efficient models can still deliver excellent results for smaller datasets like IEEE Data Port (2,141 samples), while GPT-2.0's struggles likely stem from its generative nature, which is less suited to classification tasks.

*3) Tweeteval Dataset:* RoBERTa achieved the highest accuracy (96%), outperforming all other methods, followed by BERT and XLNet with strong performances of 95% and 94%, respectively. DistilBERT achieved a moderate accuracy of 84%, while GPT-2.0 had the lowest performance at 74%. These results underscore RoBERTa's superiority on this dataset, highlighting its robustness for text classification tasks, while GPT-2.0's significantly lower accuracy reaffirms its limitations compared to classification-focused models.

TABLE V
PERFORMANCE COMPARISON OF DETECTION METHODS ACROSS DATASETS IN TERM OF EVALUATION METRICS ACCURACY (A), PRECISION (P), RECALL (R), F1 SCORE (F), ERROR RATE (E), AND INFERENCE TIME (IT). THE HIGHEST ACCURACY, PRECISION, RECALL, AND F1 SCORE, AS WELL AS THE LOWEST ERROR RATE AND INFERENCE TIME, ARE HIGHLIGHTED IN **BOLD** FOR EACH METHOD.

| Dataset | Classes | BERT | | | | | | RoBERTa | | | | | | XLNet | | | | | | DistilBERT | | | | | | GPT-2.0 | | | | | |
|---|---|---|---|---|---|---|---|---|---|---|---|---|---|---|---|---|---|---|---|---|---|---|---|---|---|---|---|---|---|---|---|
| | | A (%) | P (%) | R (%) | F (%) | E (%) | IT (sec) | A (%) | P (%) | R (%) | F (%) | E (%) | IT (sec) | A (%) | P (%) | R (%) | F (%) | E (%) | IT (sec) | A (%) | P (%) | R (%) | F (%) | E (%) | IT (sec) | (%) | P (%) | R (%) | F (%) | E (%) | IT (sec) |
| Kaggle | n = 5 | 94 | **95** | 94 | 94 | 6 | **0.053** | 93 | 94 | 93 | 93 | 7 | **0.029** | 93 | 93 | 93 | 93 | 7 | 0.397 | **92** | **91** | **92** | **91** | **8** | 0.060 | **85** | **85** | **85** | **84** | **15** | 1012 |
| IEEE Data Port | n = 5 | 83 | 83 | 83 | 83 | 17 | 0.079 | 82 | 83 | 82 | 82 | 18 | 0.063 | 85 | 86 | 85 | 85 | 15 | **0.022** | 87 | 88 | 88 | 88 | 13 | **0.001** | 76 | 80 | 76 | 77 | 24 | .017 |
| TweetEval | n = 5 | **95** | **95** | **95** | **95** | **5** | 0.095 | **96** | **95** | **96** | **95** | **4** | 0.048 | **94** | **95** | **94** | **94** | **6** | 0.432 | 84 | 78 | 84 | 81 | 16 | 0.015 | 74 | 74 | 74 | 73 | 26 | **.007** |

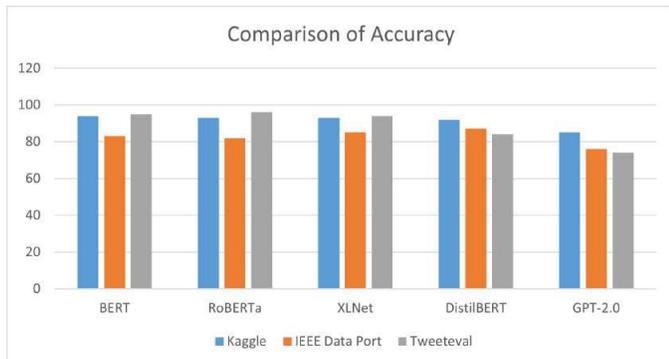

Fig. 2. Comparison of Accuracy

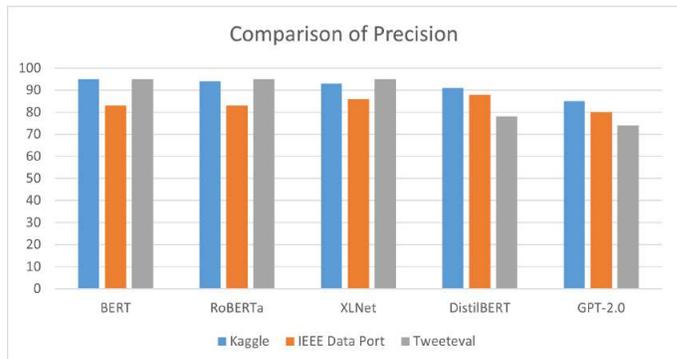

Fig. 3. Comparison of Precision

*4) Cross-Dataset Comparison:* RoBERTa and BERT consistently perform well across all three datasets, achieving top results in TweetEval (96% and 95%) and near-top results in the other datasets, while XLNet maintains competitive performance with results close to both. DistilBERT performs surprisingly well, particularly on the IEEE Data Port dataset (87%), demonstrating that smaller models can be effective for limited resources or smaller datasets. In contrast, GPT-2.0 consistently underperforms across all datasets, achieving the lowest accuracy each time, with its best result being 85% on Kaggle. These findings highlight RoBERTa and BERT as the most reliable models, while GPT-2.0's results suggest that generative models are less effective for cyberbullying detection compared to transformer-based classifiers.

The comparison highlights that RoBERTa is the most effective detection method across datasets, followed closely by BERT and XLNet. While DistilBERT performs well on smaller datasets, GPT-2.0 consistently underperforms, suggesting that generative models are less suitable for text classification tasks like cyberbullying detection. For practical deployment, RoBERTa and BERT remain the most reliable options, offering a strong balance between accuracy and computational efficiency.

### B. Comparison of Precision

*1) Kaggle Dataset:* BERT achieves the highest precision at 95%, highlighting its strong performance in correctly identifying positive cases. RoBERTa closely follows with 94%, while XLNet is slightly behind at 93%. DistilBERT shows a slight drop, achieving 91% precision. GPT-2.0 performs the worst on this dataset, with 85% precision. BERT is the most precise model on the Kaggle dataset, while GPT-2.0 lags significantly behind.

*2) IEEE Data Port Dataset:* DistilBERT achieves the highest precision at 88%, showcasing its effectiveness on the IEEE Data Port dataset, followed closely by XLNet with 86%, demonstrating strong performance. BERT and RoBERTa perform similarly, each achieving 83% precision, while GPT-2 ranks the lowest with 80% precision. Overall, DistilBERT stands out as the best model for precision, whereas GPT-2 struggles in comparison.

*3) Tweeteval Dataset:* BERT, RoBERTa, and XLNet all achieve the highest precision at 95%, showcasing equal performance on the Tweeteval dataset, while DistilBERT drops significantly to 78% precision. GPT-2 ranks the lowest with 74% precision, indicating weaker performance. Overall, BERT, RoBERTa, and XLNet excel equally in precision, whereas DistilBERT and GPT-2 underperform.

*4) Cross-Dataset Comparison:* BERT delivers consistently high precision across all datasets, particularly excelling on Kaggle and Tweeteval. RoBERTa and XLNet are equally competitive, closely matching BERT's precision performance. DistilBERT performs well on the IEEE Data Port dataset, where it outperforms other methods. However, it struggles on the Tweeteval dataset. GPT-2.0 consistently ranks the lowest in precision across all datasets, showing its limitations compared to other methods.

### C. Comparison of Recall

*1) Kaggle Dataset:* BERT, RoBERTa, and XLNet perform equally well on the Kaggle dataset, achieving recalls of 94% and 93%, respectively, while DistilBERT follows closely with a recall of 92%, remaining competitive. In contrast, GPT-2 underperforms compared to the other models, with a recall of 85%.

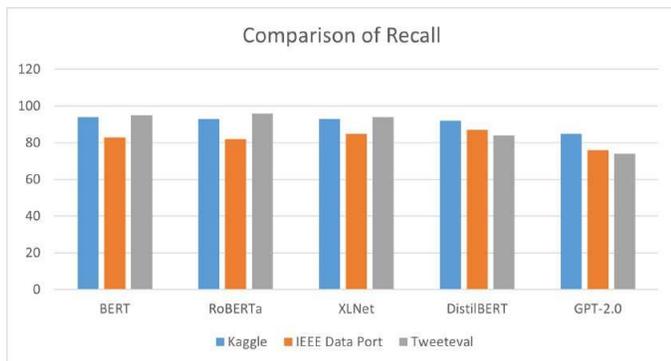

Fig. 4. Comparison of Recall

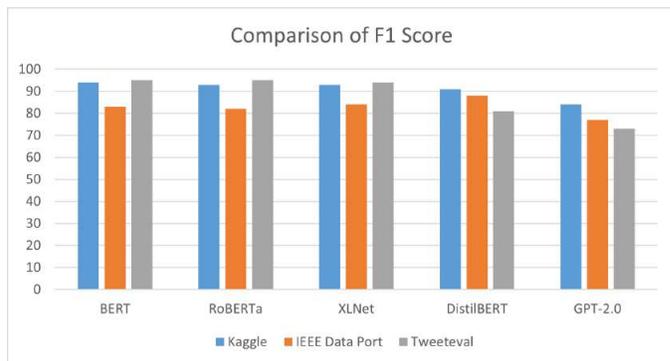

Fig. 5. Comparison of F1 score

*2) IEEE Data Port Dataset:* DistilBERT achieves the highest recall at 87%, showcasing its strong generalization ability on this dataset despite being a lighter version of BERT. XLNet follows with a recall of 85%, while BERT and RoBERTa perform similarly, each achieving 83%. GPT-2 lags significantly behind, with a recall of 76%.

*3) Tweeteval Dataset:* RoBERTa achieves the highest recall at 96%, followed closely by BERT and XLNet, both scoring 95%. DistilBERT shows a significant drop with a recall of 84%, while GPT-2 performs the worst on this dataset, achieving only 74% recall.

*4) Cross-Dataset Comparison:* RoBERTa and BERT consistently achieve high recall across all datasets, demonstrating their robustness for recall-based evaluation. XLNet also performs strongly, maintaining competitive recall scores on all datasets. DistilBERT while it performs exceptionally well on the IEEE Data Port dataset (87%), its performance declines slightly on the Kaggle and Tweeteval datasets. GPT-2.0 consistently underperforms compared to other models across all datasets, suggesting that generative models like GPT-2 may not be well-suited for tasks requiring high recall in cyberbullying detection.

RoBERTa and BERT emerge as the most reliable models for achieving high recall across all datasets. XLNet maintains competitive performance and serves as a strong alternative. DistilBERT performs exceptionally well on IEEE Data Port but shows variability on other datasets. GPT-2.0 consistently underperforms, indicating it may not be ideal for recall-focused tasks in cyberbullying detection.

### D. Comparison of F1 score

*1) Kaggle Dataset:* BERT, RoBERTa, and XLNet achieve very similar F1-Scores of 94% and 93%, respectively, indicating strong and consistent performance across these models. DistilBERT follows with an F1-Score of 91%, which is slightly lower but still competitive, while GPT-2 lags behind with the lowest F1-Score of 84%.

*2) IEEE Data Port Dataset:* DistilBERT outperforms all other models, achieving the highest F1-Score at 88%. XLNet follows with 84%, demonstrating strong performance on this dataset. BERT and RoBERTa deliver similar results, with F1-Scores of 83% and 82%, respectively, while GPT-2 ranks the lowest with an F1-Score of 77%.

*3) Tweeteval Dataset:* BERT and RoBERTa achieve the highest F1-Scores at 95%, demonstrating exceptional performance on this dataset, followed closely by XLNet with an F1-Score of 94%. DistilBERT shows a significant drop, achieving 81%, while GPT-2 performs the worst with an F1-Score of 73%.

*4) Cross-Dataset Comparison:* BERT and RoBERTa consistently achieve the highest F1-Scores across all datasets, proving their robustness and ability to balance precision and recall. XLNet performs strongly as well, showing minimal difference compared to BERT and RoBERTa. DistilBERT achieves exceptional performance on the IEEE Data Port dataset but falls behind on the Kaggle and Tweeteval datasets. This highlights that while DistilBERT is lightweight and efficient, its generalization may vary depending on the dataset. GPT-2.0 consistently achieves the lowest F1-Scores across all datasets, suggesting that it struggles to balance precision and recall for cyberbullying detection tasks.

BERT, RoBERTa, and XLNet are the top-performing models across all datasets, achieving high and consistent F1-Scores. DistilBERT performs exceptionally well on the IEEE Data Port dataset but lags behind on others. GPT-2.0 consistently underperforms, making it less suitable for tasks requiring strong F1-Score performance.

### E. Comparison of Error Rate

*1) Error Rate on the Kaggle Dataset:* DistilBERT achieves the lowest error rate at 8%, followed closely by XLNet and RoBERTa, both with an error rate of 7%. BERT has a slightly better error rate of 6%, while GPT-2 performs the worst, with the highest error rate of 15%.

*2) IEEE Data Port Dataset:* DistilBERT has the lowest error rate at 13%, followed by BERT and XLNet with comparable error rates of 17% and 15%, respectively. RoBERTa shows a similar error rate of 18%, while GPT-2 performs the worst with the highest error rate at 24%.

*3) Tweeteval Dataset:* BERT achieves the lowest error rate at 5%, closely followed by RoBERTa at 4% and XLNet at 6%. DistilBERT shows a higher error rate of 16%, while GPT-2 performs the worst, with the highest error rate of 26%.

*4) Cross-Dataset Comparison:* BERT and RoBERTa consistently perform well across datasets, achieving some of the

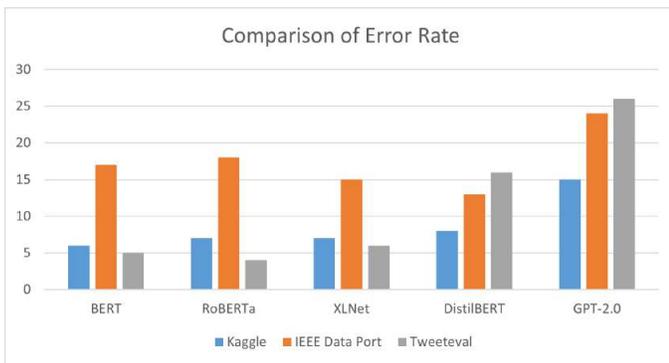

Fig. 6. Comparison of Error Rate

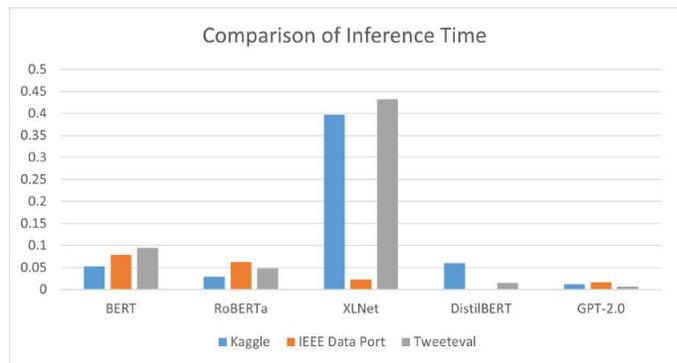

Fig. 7. Comparison of Inference Time

lowest error rates. DistilBERT excels on the Kaggle and IEEE Data Port datasets but struggles on the Tweeteval dataset. XLNet performs consistently well, though slightly higher than BERT and RoBERTa in some cases. GPT-2.0 consistently exhibits the highest error rates across all datasets, suggesting that it is not as effective for this task compared to the transformer-based classification models. The Kaggle dataset shows relatively low error rates for most models, highlighting its easier detectability. The IEEE Data Port and Tweeteval datasets present greater challenges, especially for DistilBERT and GPT-2.0.

BERT and RoBERTa emerge as the top-performing models, achieving consistently low error rates across all datasets. DistilBERT is competitive but struggles on the Tweeteval dataset. GPT-2.0 performs the worst across all datasets, consistently exhibiting the highest error rates. XLNet remains a strong contender, balancing performance across datasets.

### F. Comparison of Inference Time

*1) Kaggle Dataset:* RoBERTa achieves the fastest inference time at 0.029 seconds, followed closely by GPT-2 with 0.012 seconds, which is notably fast. BERT shows a slightly higher inference time of 0.053 seconds, while DistilBERT performs efficiently at 0.06 seconds. XLNet is the slowest model, with an inference time of 0.397 seconds.

*2) IEEE Data Port Dataset:* DistilBERT achieves the fastest inference time at 0.001 seconds, which is remarkably efficient. XLNet follows with a time of 0.022 seconds, while GPT-2 maintains a reasonable inference time of 0.017 seconds. RoBERTa and BERT have slightly slower inference times at 0.063 seconds and 0.079 seconds, respectively.

*3) Tweeteval Dataset:* GPT-2 achieves the fastest inference time at 0.007 seconds, followed by RoBERTa with 0.048 seconds. DistilBERT also shows reasonable efficiency at 0.015 seconds, while BERT has a slightly slower time of 0.095 seconds. XLNet is again the slowest model, with an inference time of 0.432 seconds.

*4) Cross-Dataset Comparison:* DistilBERT is the fastest overall on the IEEE Data Port dataset. RoBERTa achieves the fastest inference time on the Kaggle dataset. GPT-2.0 is the most efficient model on the Tweeteval dataset. XLNet consistently has the longest inference times across all datasets, making it less suitable for real-time or resource-constrained environments. RoBERTa and GPT-2.0 balance both speed and reasonable performance on most datasets, particularly on Kaggle and Tweeteval. BERT has moderate inference times, which are slower compared to RoBERTa, DistilBERT, and GPT-2.0. DistilBERT excels on the IEEE Data Port dataset but shows slightly higher times on the Kaggle and Tweeteval datasets.

GPT-2.0 and DistilBERT are the most efficient models in terms of inference time across most datasets, achieving the fastest results. RoBERTa also performs efficiently on the Kaggle dataset. XLNet is consistently the slowest model, indicating inefficiencies in inference speed. For real-time applications or situations requiring minimal inference time, DistilBERT, GPT-2.0, and RoBERTa are preferable choices.

### G. Comparison of Computational Resources

*1) Kaggle Dataset:* BERT and DistilBERT have the lowest memory usage at 35.28 MB and 1.09 MB, respectively. RoBERTa has the highest memory usage at 314 MB, followed by GPT-2 at 252.68 MB and XLNet at 212.55 MB. GPT-2 consumes the highest CPU/GPU usage at 5.3%, while BERT and XLNet have very low usage at 0.4% and 0.001%, respectively. RoBERTa and DistilBERT have moderate CPU/GPU usage at 2.4% and 0.6%. In terms of energy consumption, BERT and RoBERTa consume minimal energy at 0.000263 kWh and 0.000028 kWh, respectively, while GPT-2 and XLNet have higher energy consumption at 0.0365 kWh and 0.000884 kWh. DistilBERT is efficient with 0.0211 kWh.

*2) IEEE Data Port Dataset:* DistilBERT has the lowest memory usage at 1.2 MB, while RoBERTa consumes the most memory at 456.04 MB, followed by XLNet at 414.9 MB. GPT-2 has the highest CPU/GPU usage at 3.1%, with BERT and XLNet using minimal resources at 1.9% and 0.1%, respectively. DistilBERT and RoBERTa have moderate CPU/GPU usage at 1.3% and 1.2%. In terms of energy consumption, RoBERTa and XLNet show minimal usage at 0.000001 kWh and 0.000002 kWh, while DistilBERT and GPT-2 consume more energy at 0.0084 kWh and 0.0085 kWh, respectively. BERT uses energy at 0.0112 kWh.

*3) Tweeteval Dataset:* DistilBERT has the lowest memory usage at 100.96 MB, while RoBERTa and XLNet consume

more memory at 342.24 MB and 231.66 MB, respectively. RoBERTa has the highest CPU/GPU usage at 5.7%, followed by GPT-2 at 2.5%. DistilBERT and BERT show moderate usage at 2.6%, while XLNet uses very little at 0.001%. In terms of energy consumption, GPT-2 exhibits significant usage at 0.0549 kWh, while RoBERTa and XLNet have moderate energy consumption at 0.000118 kWh and 0.000964 kWh, respectively. BERT and DistilBERT use very little energy at 0.000507 kWh and 0.000109 kWh.

*4) Cross-Dataset Comparison:* DistilBERT consistently uses the least memory across all datasets, making it ideal for low-memory environments. RoBERTa and GPT-2.0 require significant memory resources. GPT-2.0 and RoBERTa have the highest CPU/GPU usage, indicating higher computational costs. XLNet uses the least CPU/GPU resources but sacrifices efficiency in other areas. DistilBERT and RoBERTa are the most energy-efficient models overall. GPT-2.0 consumes the most energy, especially for the Tweeteval dataset.

DistilBERT is the most resource-efficient model overall, with the lowest memory usage and moderate CPU/GPU usage and energy consumption. GPT-2.0 is resource-intensive in terms of CPU/GPU usage and energy consumption, despite performing well in other areas. RoBERTa and XLNet trade-offs include high memory requirements but offer energy efficiency. For environments with limited computational resources, DistilBERT is the preferred choice. For high-performance systems where energy consumption is less of a concern, GPT-2.0 and RoBERTa can be considered.

RoBERTa emerges as the best detection method in terms of performance, achieving the highest accuracy, precision, and F1-scores across all datasets (Kaggle, IEEE Data Port, and TweetEval), although with higher computational costs. XLNet closely follows RoBERTa in performance but is more resource-intensive. DistilBERT stands out as the most efficient model, offering the fastest inference time, minimal memory usage, lowest CPU/GPU usage, and energy consumption, making it ideal for resource-constrained environments. BERT strikes a balance between performance, time efficiency, and resource usage, making it a versatile option. While GPT-2.0 shows moderate performance, it struggles with higher resource consumption and slower inference times. Overall, RoBERTa is the best choice for high performance, DistilBERT excels in efficiency, and BERT offers a balanced trade-off between the two.

### H. Performance Insights

BERT achieves high performance due to its bidirectional transformer architecture, enabling it to capture contextual information both before and after a token. This enhances its ability to classify text accurately, reducing misclassifications of cyberbullying, and its pre-training on large datasets ensures robust generalization across diverse datasets. RoBERTa demonstrates even higher performance by eliminating the Next Sentence Prediction (NSP) task, training on larger datasets with extended sequences, and benefiting from fine-tuning and hyperparameter optimization. In contrast, XLNet shows moderate performance, as it does not leverage bidirectional context as effectively as BERT or RoBERTa and is more sensitive to data imbalance and sequence length variability. DistilBERT, with its reduced layers, offers faster but slightly less comprehensive predictions, resulting in moderate performance. GPT-2, relying on a unidirectional transformer architecture, performs poorly in comparison, as it cannot capture context as effectively as bidirectional models like BERT and RoBERTa.

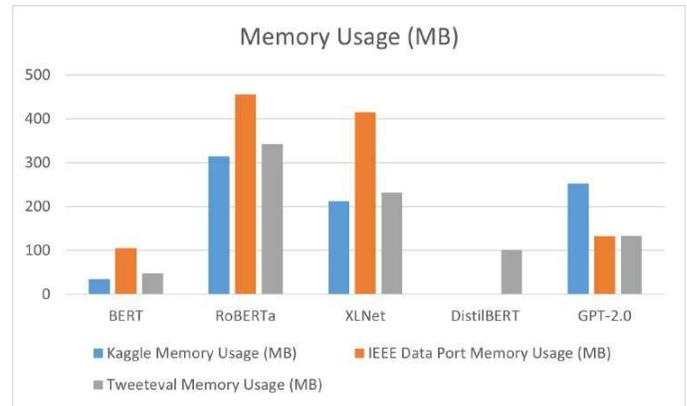

Fig. 8. Comparison of Memory Usage

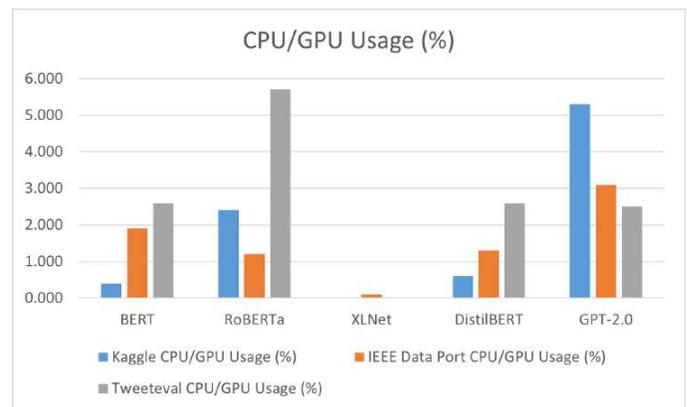

Fig. 9. Comparison of CPU/GPU Usage

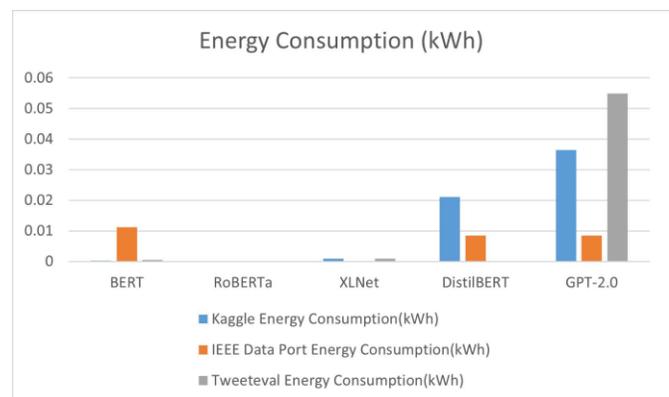

Fig. 10. Comparison of Energy Consumption

TABLE VI
COMPUTATIONAL RESOURCE USAGE COMPARISON ACROSS DATASETS AND METHODS. THE LOWEST RAM USAGE, CPU USAGE, AND ENERGY CONSUMPTION ARE HIGHLIGHTED IN **BOLD** FOR EACH METHOD.

| Dataset | Classes | Parameter | BERT | RoBERTa | XLNet | DistilBERT | GPT-2.0 |
|---------|---------|-----------|------|---------|-------|------------|---------|
| Kaggle | 5 | Memory Usage (MB) | **35.28** | 314.00 | 212.55 | **1.09** | 252.68 |
| | | CPU/GPU Usage (%) | **0.400** | 2.400 | **0.001** | **0.600** | 5.300 |
| | | Energy Consumption (kWh) | **0.000263** | 0.000028 | 0.000884 | 0.021100 | 0.036500 |
| IEEE Data Port | 5 | Memory Usage (MB) | 105.23 | 456.04 | 414.90 | 1.20 | **132.04** |
| | | CPU/GPU Usage (%) | 1.90 | **1.20** | 0.10 | 1.30 | 3.10 |
| | | Energy Consumption (kWh) | 0.011200 | **0.000001** | **0.000002** | 0.008400 | **0.008500** |
| TweetEval | 5 | Memory Usage (MB) | 47.58 | 342.24 | 231.66 | 100.96 | 133.17 |
| | | CPU/GPU Usage (%) | 2.6000 | 5.7000 | 0.0011 | 2.6000 | **2.5000** |
| | | Energy Consumption (kWh) | 0.000507 | 0.000118 | 0.000964 | **0.000109** | 0.054900 |

### I. Time Efficiency Insights

BERT demonstrates moderate time efficiency due to its deep architecture with twelve layers in the base model, making it computationally intensive; however, modern framework optimizations help maintain reasonable inference times. RoBERTa, designed for maximum performance, exhibits low time efficiency, as its deeper training and lack of inference optimizations result in slower processing. XLNet shows very low time efficiency, with its autoregressive pre-training and deeper architecture significantly slowing down inference compared to BERT and DistilBERT. DistilBERT, with fewer layers and parameters, achieves high time efficiency, making it suitable for real-time applications. In contrast, GPT-2 suffers from very low time efficiency due to additional overheads that result in prolonged inference times.

### J. Computational Resources Usage Insights

BERT has moderate resource usage due to its relatively large number of parameters, requiring moderate memory and computational power, though it is more optimized than XLNet and RoBERTa. RoBERTa exhibits very high resource usage, with its extended pre-training and larger data consumption making it highly resource-intensive, resulting in a larger memory footprint and greater energy consumption. XLNet also has very high resource usage, as its autoregressive modeling and large architecture demand significant RAM and energy, making it less suitable for resource-constrained environments. DistilBERT, on the other hand, has low resource usage, as its reduced number of parameters significantly lowers RAM requirements and energy consumption while maintaining reasonable performance. GPT-2, with one of the highest parameter counts, has very high resource usage, requiring substantial memory, CPU/GPU power, and energy.

### K. Recommendations on Detection Methods and Datasets

To select the best model for balanced performance and efficiency across all datasets, it is essential to prioritize models that maintain high accuracy while being computationally efficient. Models like DistilBERT or optimized versions of BERT and RoBERTa are ideal for real-time or resource-constrained applications. Avoid outdated generative models like GPT versions one and two due to inefficiency and low performance, and instead leverage updated versions such as GPT-3 or GPT-4, which offer improved efficiency and performance.

Optimize model inference time and resource usage by applying techniques like model pruning, quantization, and knowledge distillation. These methods reduce model complexity while retaining performance, making them suitable for lightweight deployment in resource-constrained environments.

To improve dataset quality, focus on enhancing source, size, scope, diversity, and generalizability. Collaborate with related institutions, such as those in mental health and psychology, to access valuable insights and data directly related to cyberbullying. Emphasize better annotation practices and address class imbalances to reduce sensitivity to specific datasets. Ensure that datasets reflect the dynamic nature of cyberbullying by considering trends, scenarios, and new generative AI techniques for realistic synthetic data creation.

Real-world datasets from social media platforms should be prioritized for realism, aiding in the development of high-quality datasets for the cyberbullying detection domain. Additionally, combining multiple optimization techniques can yield lightweight yet high-performing models, tailored to text classification tasks in this domain. Finally, prioritize energy-efficient models to ensure sustainability in energy-sensitive deployment environments.

Table VII provides a summary of the strengths, limitations, and evaluation criteria of existing detection methods. While studies have shown promising results, further improvement is needed in model effectiveness, time efficiency, computational costs, generalizability, and scalability. Advancing these models with ongoing development and advanced techniques can lead to robust, reliable, real-time tools with minimal computational costs and enhanced scalability, helping to mitigate the harmful effects of cyberbullying on communities and making social media platforms safer and more reliable.

TABLE VII
COMPARISON OF METHODS, STRENGTHS, LIMITATIONS, AND EVALUTION CRITERIA

| Method | Strength | Limitation | Evaluation Level | | |
|--------|----------|------------|------|------|------|
| | | | PA | TE | CR |
| BERT | Strong performance across social media platforms | Heavily dependent on labeled data, Relatively high inference time and computational resource usage | 2 | 5 | 2 |
| RoBERTa | Achieves competitive performance | Less effective for small datasets, Moderately expensive in terms of computation resources | 1 | 4 | 5 |
| XLNet | Achieve high Performance and faster inference times | Flexibility limitation, expensive in computation resources | 3 | 3 | 3 |
| DistilBERT | Significantly reduced inference time and computational resource usage, achieve high performance | Slit drop of performance | 4 | 1 | 1 |
| GPT-2.0 | Highly robust, moderate computational efficiency | Shows lower performance | 5 | 2 | 4 |

**Key:**
**"1"** indicates the highest level, **"2"** indicates a high level, **"3"** indicates a moderate level, **"4"** indicates a low level, **"5"** indicates the lowest level.
**"PA"**, **"TE"**, **""CR**" stands for Performance Analysis, Time Efficiency, and Computational Resources respectively.

## VI. FUTURE WORK AND CONCLUSION

This section highlights the key areas for further exploration to advance the field of Text Classification. It also summarizes the findings of the study and examines their implications for strengthening Text Classification in the domain of Cyberbullying Detection.

### A. Future Work

*1) Comparative Analysis of Mitigation Approaches for Cyberbullying Detection and Prevention:* Future studies will conduct a comparative analysis by adapting and refining mitigation mechanisms from general text classification, assessing their ability to accurately identify harmful text on social media platforms in the context of cyberbullying classification for real-time detection. The study will explore how effective the mitigation methods adapted from general text classification are in accurately identifying harmful text on social media platforms within the cyberbullying detection domain. This study will help identify the best mitigation approaches for blocking cyberbullying on social media platforms and creating a safer community.

*2) Development of a Unified Novel System for Text Classification of Cyberbullying Instances:* Future research will focus on developing a unified, novel system for text classification of cyberbullying instances. This approach can then be extended to other low-resource languages, using Swahili as a case study, thereby enhancing the overall robustness and reliability of social media platforms. The research will explore how an efficient, novel system can be developed to accurately detect harmful content in both English and Swahili texts in instances of cyberbullying on social media platforms.

Investigating these future research directions will significantly advance detection, monitoring, and mitigation methods for text classification in the cyberbullying domain. By conducting a comparative analysis of mitigation approaches and developing a system that classifies harmful text in both English and Swahili, this research will enhance the effectiveness of cyberbullying detection. This holistic approach will result in more effective solutions, ultimately enhancing the safety and dependability of social media platforms.

### B. Conclusion

The study emphasizes the effectiveness of general text classification techniques in adapting and applying them to identify cyberbullying instances on social media platforms within the cyberbullying detection domain. Through the evaluation of detection methods: BERT, RoBERTa, XLNet, DistilBERT, and GPT-2.0, various performance trade-offs were observed. RoBERTa emerged as the top-performing method, achieving the highest accuracy, precision, and F1-scores across all datasets (Kaggle, IEEE Data Port, and TweetEval), though it comes with higher computational costs. XLNet closely follows RoBERTa in terms of performance but is more resource-intensive. DistilBERT stands out for its efficiency, offering the fastest inference time, minimal memory usage, and the lowest CPU/GPU usage and energy consumption, making it ideal for resource-constrained environments. BERT strikes a balance between performance, time efficiency, and resource consumption, making it a versatile option. While GPT-2.0 demonstrates moderate performance, it struggles with higher resource consumption and slower inference times. Overall, RoBERTa is the best choice for high performance, DistilBERT excels in efficiency, and BERT provides a balanced trade-off between the two.

Beyond these perceptions, there is a critical need to develop detection methods specifically designed for the text classification of cyberbullying instances on social media platforms. The methods evaluated in this study were adapted from the general text classification domain, resulting in only moderate detection rates, particularly when addressing complex forms of cyberbullying. Although influencing existing advancements in cyberbullying detection has been valuable, this adaptation has highlighted the limitations of these methods in effectively tackling the unique challenges inherent in text classification within the cyberbullying detection domain when varying different social media platforms.

To guarantee the safety and reliability of social media platforms for the internet users, it is crucial to prioritize the development of specialized detection methods that effectively tackle the complexities of text classification within the cyberbullying domain. By emphasizing these modified approaches, the robustness and effectiveness of cyberbullying detection methods can be significantly enhanced, eventually promoting safer and more supportive online communities.

Source Code and Dataset Availability Statement: The source code and datasets used in this study are available at:https://github.com/adamu1/Assessing-Text-Classification-Methods-for-Cyberbullying-Detection-on-Social-Media-Platforms

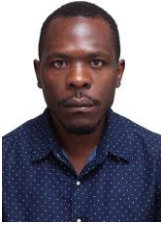

**Adamu        Gaston        Philipo** (d202361019@xs.ustb.edu.cn) received his Bachelor of Science degree in Electronics and Telecommunications Engineering from the University of Dodoma (UDOM), Dodoma, Tanzania, in 2017 and his Master of Science degree in Cybersecurity and Digital Forensics from Coventry University, Coventry, United Kingdom, in 2021. He is currently pursuing a Ph.D. in Computer Science and Technology at the School of Computer and Communication Engineering, University of Science and Technology Beijing (USTB), Beijing, China. His current research interests include Cybersecurity, and Cyberbullying Detection.

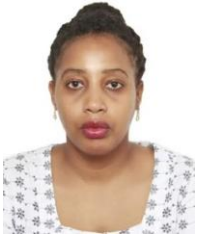

**Doreen        Sebastian        Sarwatt** (d202161035@xs.ustb.edu.cn) received her Bachelor of Science degree in Computer Engineering and IT from the University of Dar es Salaam (UDSM), Dar es Salaam, Tanzania, in 2009 and her Master of Science degree in Computer Science from the University of Dar es Salaam (UDSM), Dar es Salaam, Tanzania, in 2015. She is currently pursuing a Ph.D. in Computer Science and Technology at the School of Computer and Communication Engineering, University of Science and Technology Beijing, China. Her research interests include Cybersecurity, Cyberbullying Detection, Autonomous Vehicles, Adversarial Attacks, Computer Vision, and Image Processing.

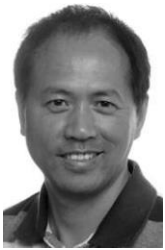

**Jianguo Ding** (jianguo.ding@bth.se) received his Doctorate in Engineering (Dr.-Ing.) from the faculty of mathematics and computer science at the University of Hagen, Germany. He is currently an Associate Professor at the Department of Computer Science, Blekinge Institute of Technology, Sweden. His research interests include cybersecurity, critical infrastructure protection, intelligent technologies, blockchain, distributed systems management and control, and serious games. He is a Senior Member of the IEEE (SM'11) and a Senior Member of the ACM (SM'20).

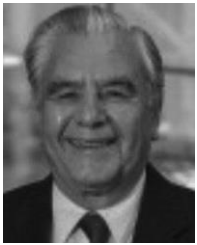

**Mahmoud Daneshmand** (Life Senior Member, IEEE) received BS and M.S in Mathematics from the University of Tehran, Iran; M.S and Ph.D degrees in Statistics from the University of California at Berkeley, CA, USA, in 1973 and 1976, respectively. He is currently an Industry Professor with the Department of Business Intelligence and Analytics as well as Department of Computer Science at Stevens Institute of Technology, USA. He has more than 40 years of Industry and University experience as Professor, Researcher, Assistant Chief Scientist, Executive Director, Distinguished Member of Technical Staff, Technology Leader, Chairman of Department, and Dean of School at: Bell Laboratories; AT&T Shannon Labs Research; University of California, Berkeley; University of Texas, Austin; Sharif University of Science and Technology; University of Tehran; National University of Iran; New York University; and Stevens Institute of Technology. He has published more than 300 journal and conference papers; authored/coauthored three books.

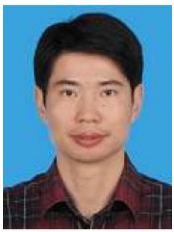

**Huansheng Ning** (ninghuansheng@ustb.edu.cn) received the B.S. degree from Anhui University, Hefei, China, in 1996 and the Ph.D. degree from Beihang University, Beijing, China, in 2001. He is currently a Professor with the School of Computer and Communication Engineering, University of Science and Technology Beijing, China, and the founder and the principal at Beijing Cyberspace International Science and Technology Cooperation Base. His current research interests include IoT, General Cyberspace and Materverse, Smart Education, Cyber-syndrome and Cyber-Health.